\documentclass[letterpaper, 10 pt, conference]{ieeeconf}
\IEEEoverridecommandlockouts
\usepackage{gensymb}
\usepackage{graphicx}
\usepackage{amsmath}
\usepackage{cite}
\usepackage{xcolor}
\title{\LARGE \bf
Ground Perturbation Detection via Lower-Limb Kinematic States During Locomotion}

\author{Maria T. Tagliaferri$^{1}$, Leonardo Campeggi$^{1}$, Owen N. Beck$^{2}$, and Inseung Kang$^{1}$
\thanks{$^{1}$M. T. Tagliaferri, L. Campeggi, and I. Kang are with the Department of Mechanical Engineering, Carnegie Mellon University, Pittsburgh, PA, 15213 USA (e-mail: {\tt\footnotesize mtagliaf@cmu.edu}).}
\thanks{$^{2}$O. N. Beck is with the Department of Kinesiology and Health Education, The University of Texas at Austin, Austin, TX, 78712 USA.}
}

\begin{document}

\maketitle
\thispagestyle{empty}
\pagestyle{empty}

\begin{abstract}
Falls during daily ambulation activities are a leading cause of injury in older adults due to delayed physiological responses to disturbances of balance. Lower-limb exoskeletons have the potential to mitigate fall incidents by detecting and reacting to perturbations before the user. Although commonly used, the standard metric for perturbation detection, whole-body angular momentum, is poorly suited for exoskeleton applications due to computational delays and additional tunings. To address this, we developed a novel ground perturbation detector using lower-limb kinematic states during locomotion. To identify perturbations, we tracked deviations in the kinematic states from their nominal steady-state trajectories. Using a data-driven approach, we further optimized our detector with an open-source ground perturbation biomechanics dataset. A pilot experimental validation with five able-bodied subjects demonstrated that our model distinguished perturbed from unperturbed gait cycles with 98.8\% accuracy and only a delay of 23.1\% within the gait cycle, outperforming the benchmark by 47.7\% in detection accuracy. The results of our study offer exciting promise for our detector and its potential utility to enhance the controllability of robotic assistive exoskeletons.
\end{abstract}


\section{Introduction}
Dynamic postural control during locomotion is regulated by anticipatory and reactive neuromuscular mechanisms that work to maintain stability \cite{bruijn2013locomotion_stability}. During daily ambulation tasks, these physiological systems handle occasional perturbations (i.e., sudden changes in walking stability) caused by factors such as unexpected changes in environmental contexts \cite{bruijn2018control,JKPT2021}. Maintaining stability in response to disturbances requires the rapid integration of sensory inputs and motor outputs to execute gait pattern adjustments, reducing the risk of falling \cite{hardwick2022age,VANDENBOGERT2002199}. Fall prevention strategies for older adults must compensate for delayed reaction times by detecting disturbances and initiating gait stabilization adjustments faster than the physiological response \cite{beck2023exoskeletons,mccrum2019trip}. Lower-limb exoskeletons designed for balance correction offer a promising solution to achieve this type of fall prevention \cite{Kang2021,10452751,doi:10.1126/scirobotics.adi8852}. However, progress is limited by the lack of consensus on a standard metric for detecting ground perturbations during locomotion \cite{Shimada2003}, 
\cite{curtze2024notes}. 

A simple approach to detect walking perturbations is to identify when the person's body center of mass (COM) moves outside the area bounded by both feet \cite{curtze2024notes,hof2008extrapolated}. However, this detection mechanism is unsuitable for exoskeleton control due to the challenge in defining consistent boundaries throughout the gait cycle (e.g., single vs. double stance) during locomotion \cite{of2005condition}. Whole-body angular momentum (WBAM) is commonly proposed as an alternative metric to detect perturbations during walking \cite{shokouhi2024recovering}. WBAM represents the combined rotational momentum of all body segments around the person's COM. Humans tightly regulate WBAM during steady-state walking; therefore, perturbations can be identified by observing a sudden increase in WBAM \cite{c9cf05b002694c58b5e09859532a20b9,ZHANG2024103179}. 

While this metric is effective for perturbation detection in retrospective gait analyses, it is not well suited for real-time control of exoskeletons \cite{shokouhi2024recovering}. Calculating WBAM involves complex kinematic analyses to determine the momentum of each limb segment in the sagittal and frontal planes. This requires substantial time and computational resources, leading to delays when monitoring the instantaneous value of WBAM \cite{ZHANG2024103179}. Another drawback is that steady-state WBAM varies significantly between walking conditions (e.g., incline walking) and subjects. Therefore, WBAM must be normalized by each subject's walking speed, body mass, and body height to maintain its accuracy \cite{LIU2023103944}. Thus, in the context of perturbation detection for assistive exoskeletons, a large amount of additional gait data may be required to calibrate a WBAM perturbation threshold across subjects and walking speeds\cite{wang2024balance}.

Direct tracking of lower-limb kinematic states can be a promising solution to ground perturbation detection for exoskeleton control \cite{bruijn2013locomotion_stability}. The key anatomical points involved in walking balance, such as the foot and COM, exhibit predictable patterns during stable locomotion. Deviations in the position and velocity of these points, relative to their steady-state trajectories, serve as clear indicators of disturbances during locomotion. These deviations can be quantified with minimal time and computation using data from motion capture systems or wearable sensing \cite{KARAKASIS2021110849}.

In this paper, we propose a novel ground perturbation detector during level-ground walking. Our model tracks 16 lower-limb kinematic states in a local coordinate system and quantifies the variation of each state from its value during steady-state walking. For our detector, we use a single threshold value to identify the perturbation onset timing across subjects and perturbation types. We evaluated our model's detection performance in trip- and slip-type ground perturbations. These perturbation scenarios were selected because they are the most common cause of falls in aging gait during walking \cite{Shimada2003}. We hypothesize that a ground perturbation detector using lower-limb kinematic states will outperform a WBAM-based model by achieving faster and more accurate detection. Our underlying rationale is that the kinematic states are highly sensitive to gait disturbances, allowing efficient and accurate tracking in real-time with minimal computation \cite{Lugris2024,JIANG2024103184}. The development of an improved walking perturbation detector will directly support advancements in balance-assistive exoskeletons and potentially reduce the risk of falls in aging adults.

\section{Methods}
\subsection{Open-Source Gait Perturbation Dataset}
We utilized an open-source gait biomechanics dataset to investigate how individuals respond to ground translation perturbations during locomotion \cite{leestma2023linking}. This dataset contained motion capture marker data, ground reaction forces, and whole-body angular momentum for 11 able-bodied subjects (4 females, age of 24.5$\pm$3.4 years, height of 175.1$\pm$7.2 cm, leg length of 91.3$\pm$5.1 cm, and body mass of 73.3$\pm$11.0 kg). The trial conditions included three sessions of 96 unique ground translation perturbations. The perturbations were varied in eight directions with three distinct magnitudes (i.e., ground translations of 5, 10, and 15 cm) using a 6 degrees of freedom actuated platform. Ground perturbations occurred at varying points within the gait cycle: 1) 50\% of the double stance phase and 2) 25\%, 50\%, and 75\% of the single stance phase. Gait phase was varied because it directly impacts subject's response to perturbations \cite{leestma2023linking}. Trials in which subjects lost force plate contact during a perturbation response, such as a jump response, were omitted from our analysis

\subsection{Tracking Kinematic States in a Local Coordinate System}
\begin{figure}[t!]
    \centering
    \includegraphics[width=0.8\linewidth]{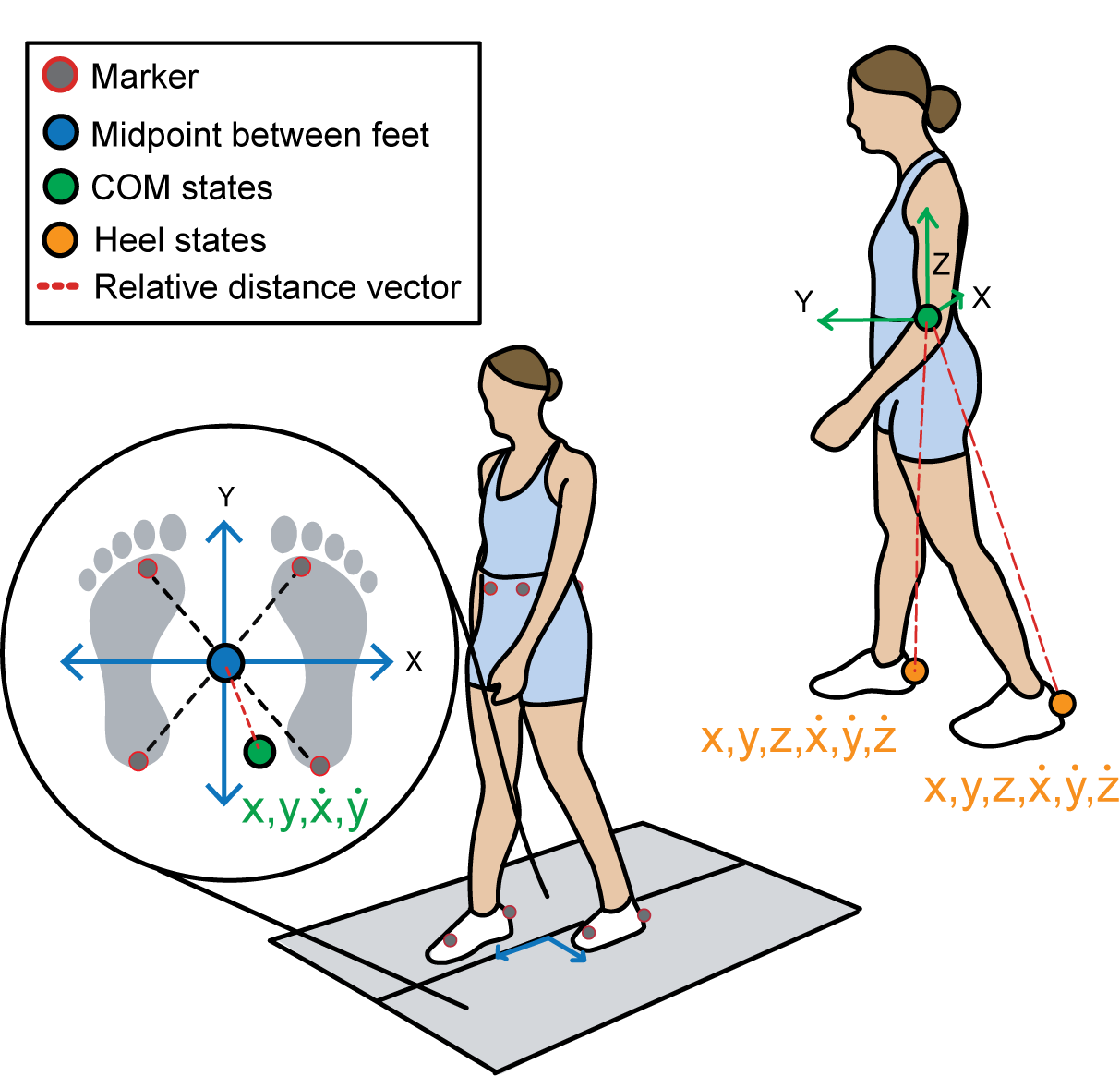}
    \caption{Kinematic state-based perturbation detection. The relative distance between the body's COM and each foot was calculated to track the kinematic state (right). Eight relative position states and eight velocity states were tracked. A local coordinate system was established at the estimated midpoint between the feet in the horizontal plane. The COM position was then projected into this coordinate system, with a vector (red dashed lines) tracking the relative distance between the COM and the feet over time (left).}
    \label{fig:fig1}
    \end{figure}

Humans tightly regulate the relative distance between the COM and the foot, as well as the velocity of each, to maintain stability during locomotion \cite{REDFERN19941339,Bagce2013}. Therefore, we tracked the relative position and velocity of the COM and both feet to detect the perturbation onset. A local coordinate system was chosen to minimize the effect of natural global positional drift during walking and to automatically scale the tracked kinematic states to the anthropomorphic dimensions of each subject (Fig. \ref{fig:fig1}). 

\begin{figure}[t!]
    \centering
    \includegraphics[width=1\linewidth]{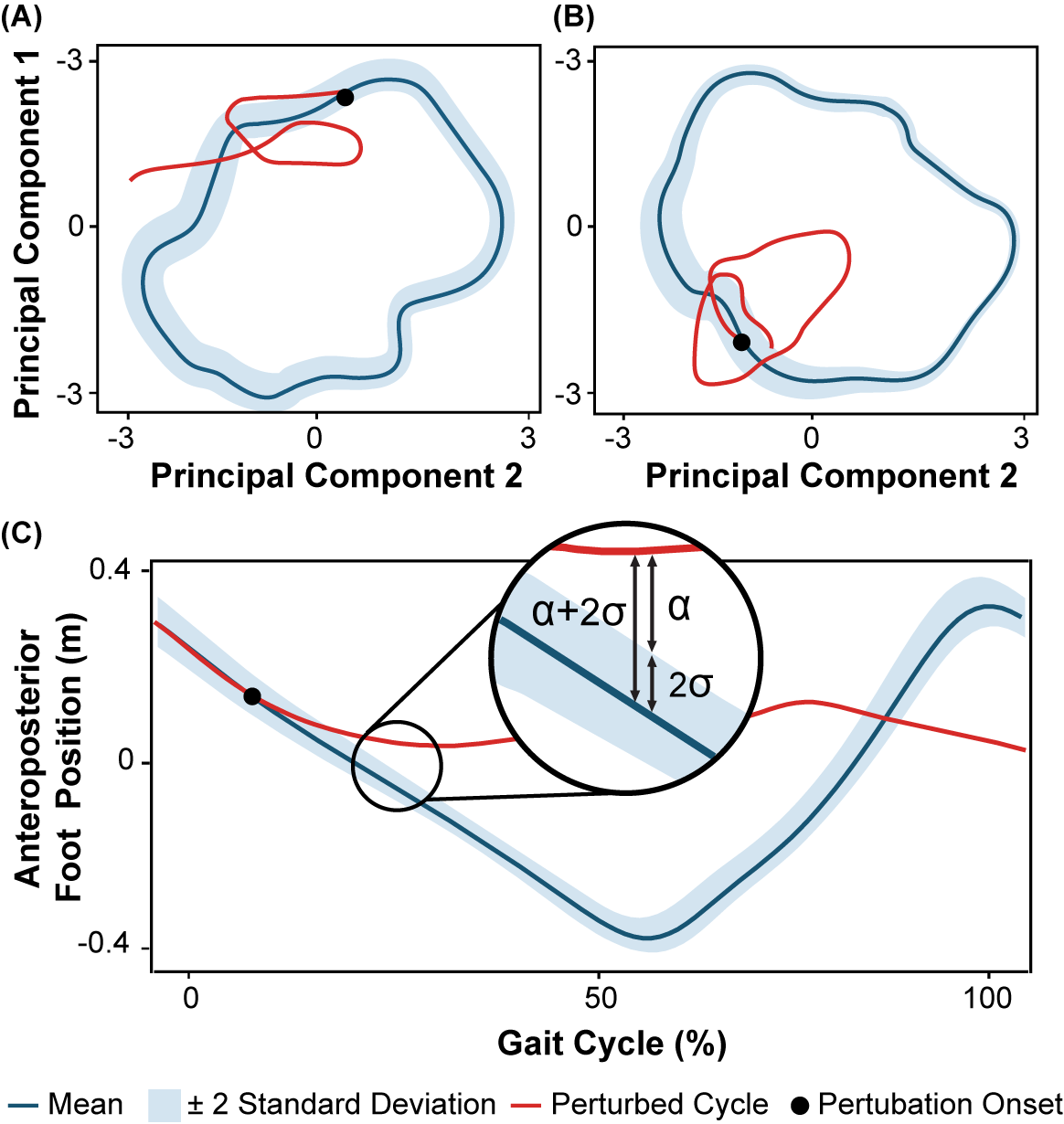}
    \caption{(A) Principal Component Analysis for a single representative trial of an (A) anteroposterior and (B) mediolateral perturbation. For both cases, perturbations with a magnitude of 5 cm occurred during the double stance phase. (C) Visual representation of state variance calculation (Eqs. \ref{eq:eq1} and 2).  \(\alpha\) is the Euclidean distance between the current state value and the edge of 2 standard deviations (\(\sigma\)) from the mean.}
    \label{fig:fig2}
    \end{figure}

To track the state of each foot, we used the subject's heel markers. The heel position was calculated within a coordinate system centered at the global COM in 3 axes (Fig. \ref{fig:fig1}). The global COM position was calculated as the average of the positions of four pelvic markers (right/left anterior/posterior superior iliac spine markers). The relative COM position was determined by projecting the global COM vertically downward onto a coordinate system centered at the instantaneous midpoint between the feet in the horizontal plane (Fig. \ref{fig:fig1}). The vertical axis was omitted because the vertical position of the COM is minimally sensitive to ground translation perturbations \cite{REDFERN19941339,curtze2024notes}. The velocity states were calculated by differentiating the global position over time. A dimensionality reduction technique using Principal Component Analysis (PCA) was implemented. This analysis provided a simplified visual inspection of the overall state response to perturbations (Fig. \ref{fig:fig2}). It also provided qualitative validation of our prediction that the selected states would exhibit rapid and distinct deviations from steady-state behavior following a perturbation. Two different perturbation trials from the open-source dataset, one in the mediolateral direction (Fig. \ref{fig:fig2}A) and one in the anteroposterior direction (Fig. \ref{fig:fig2}B), were chosen for a visual evaluation.

\subsection{Quantifying Kinematic State Variance}
The mean and standard deviation of each state were calculated as a function of the gait cycle using state values from 10 steady-state walking cycles. We chose 10 cycles as a conservative value for our initial development of the approach, however this number can likely be reduced to 2-3 cycles while still maintaining performance in future analysis \cite{Kim2024}. For subsequent gait cycles, the current value of each state was compared to the mean and standard deviation to quantify variance using Eq. \ref{eq:eq1},
\begin{equation}
\alpha_i =
\begin{aligned}
&\begin{cases}
0, & \text{if } |\bar{x}_i - x_i|\leq 2\sigma_i, \\
x_i - (\bar{x}_i \pm 2\sigma_i), & \text{otherwise}.
\end{cases}
\end{aligned}
\label{eq:eq1}
\end{equation}

where \(\bar{x}\) and \(\sigma\) represent the mean and standard deviation of the current state at a given time point, \(\alpha\) denotes the deviation of the individual state from its natural range of variance, and $i$ indicates the current time step. During walking, \(\alpha\) was assigned a value of zero if the current state value was within $\pm$2 standard deviations of the mean. Otherwise, \(\alpha\) was calculated as the euclidean distance from the current state value to 2 standard deviation above or below the mean (Fig. \ref{fig:fig2}C). A value of \(\alpha\) close to zero indicated the subject's kinematic states are within a stable basin, corresponding to steady-state walking.

Next, \(\alpha\) was used to calculate an instantaneous ratio of current to natural state variance (Fig. \ref{fig:fig2}C). This ratio ensured that a state with a lower natural variance contributed more significantly to the total stability deviation value and prevented a state with a large natural variance from disproportionally skewing the total value. The standard deviation of each state was normalized using the coefficient of variation formula, Eq. \ref{eq:eq2a},

\begin{equation}
C_{i}= \frac{\sigma_i}{\bar{x}_i} \tag{2a}
\label{eq:eq2a}
\end{equation}

to account for the differences in unit and scale between states \cite{jalilibal2021monitoring}. \( C \) represents the coefficient of variation. The ratio for each state was then computed and linearly aggregated using Eq. \ref{eq:eq2b},
 
 \begin{equation}
\varphi_{i} = \frac{1}{n} \sum_{j=1}^{n} \frac{\alpha_i}{2C_{i} + \alpha_i} \tag{2b}
\label{eq:eq2b}
\end{equation}

where \(\varphi\) represents a single stability deviation value that characterizes the subject's perturbation state and $n$ is the total number of tracked kinematic states. By monitoring the single variable, \(\varphi\), we determined whether a subject was perturbed (i.e., deviating from their natural variance) by selecting an appropriate threshold value.

\subsection{Threshold Optimization}
To calculate the optimal threshold value for \(\varphi\), which denotes a perturbation onset, we conducted a parameter sweep using the open-source perturbation dataset \cite{leestma2023linking}. Values increasing by a step size of .001 from 0.5 to 4 were used as a threshold for perturbation detection. Detection performance was evaluated on accuracy and delay at each threshold. We determined accuracy by calculating the number of gait cycles that were correctly labeled as perturbed or unperturbed. We defined delay as the percentage of gait phase that elapsed between the onset of the perturbation and the time when \(\varphi\) crossed the detection threshold. For the parameter sweep, we rewarded threshold values that minimized false detections (false positives) and undetected perturbations (false negatives), aligning with the target application of our approach for lower-limb exoskeleton control. Incorrect exoskeleton stability assistance, whether unnecessarily initiated due to false detection of a perturbation or omitted due to an undetected perturbation, carries a significant risk of user injury \cite{10479575}. 

\subsection{Baseline Benchmark Implementation}
A perturbation detection model using WBAM was implemented and evaluated as a baseline comparison for our model \cite{shokouhi2024recovering, martelli2011detecting}. We replicated this method using the instantaneous sagittal and frontal WBAM values provided in the open-source perturbation dataset. For each trial, the mean and standard deviation of WBAM were calculated using data from the 3-5 steady-state gait cycles that preceded the induced perturbation. A detection threshold of $\pm$4 standard deviations was chosen according to the recommendations of a previous literature on WBAM \cite{martelli2011detecting}. Perturbation detection occurred when the instantaneous WBAM value exceeded this threshold. We evaluated the performance of this model on both the detection accuracy and the delay.

\subsection{Pilot Human Subject Experiment}
\begin{figure}
    \centering
    \includegraphics[width=1\linewidth]{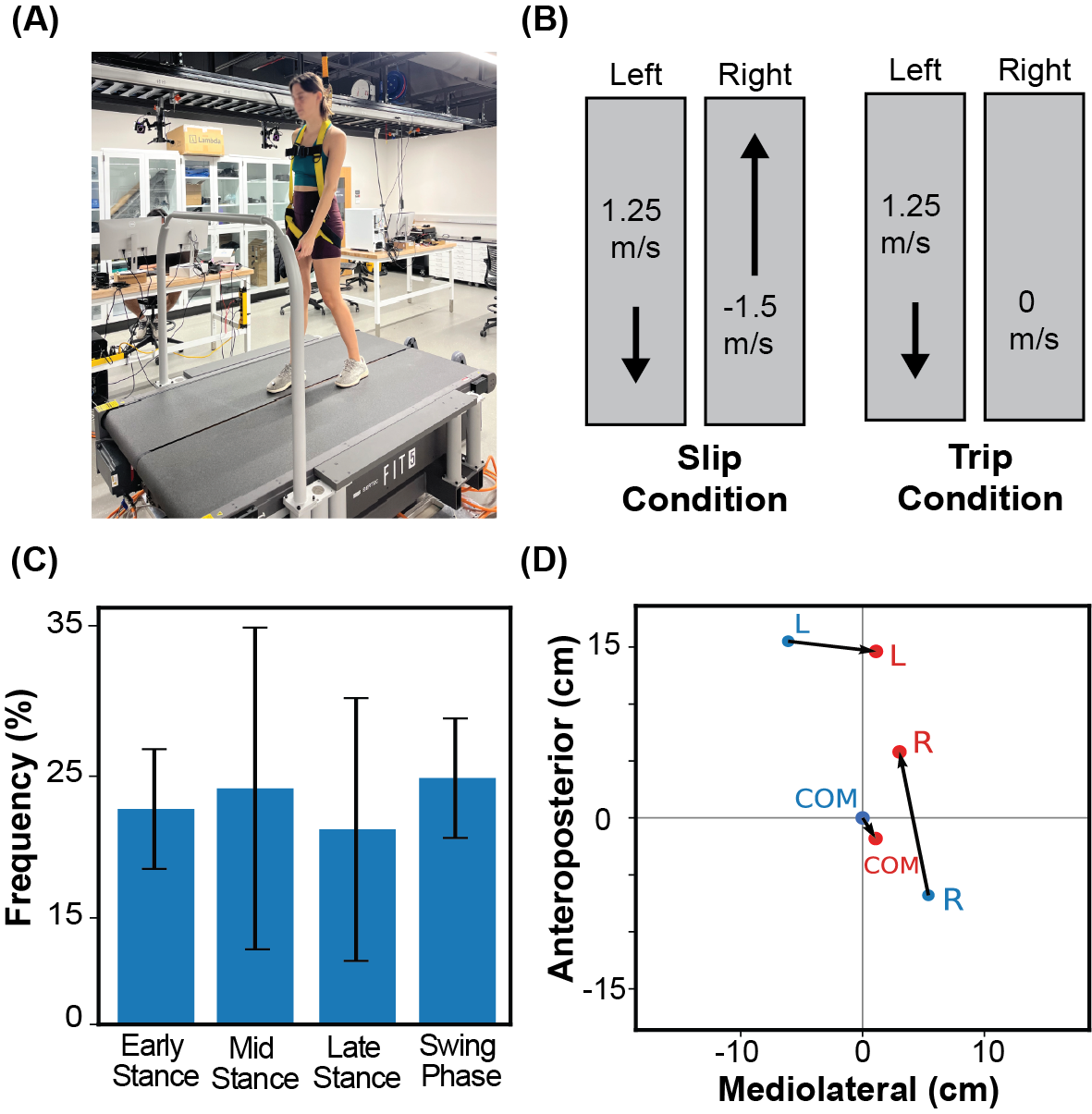}
    \caption{(A) Experimental setup. (B) Split-belt treadmill protocol for trip- and slip-type perturbations. (C) Perturbation timing: early stance (0-10\% and 90-100\% of the gait cycle), mid stance (10-30\% of the gait cycle), late stance (30-50\% of the gait cycle), and swing phase (50-90\% of the gait cycle) of the right leg. Error bars indicate $\pm$1 standard deviation. (D) Visual illustration of tracking of positional states. Blue dots represent expected positional states and red dots show actual states values during a perturbation. Black arrows indicate a distance vector between expected and actual states.} 
    \label{fig:fig3}
\end{figure}

Five participants (age of 24.2$\pm$2.1 years) were exposed to a total of 20 perturbations, divided across two sessions of 10 walking trials each. Five trip-type perturbations and five slip-type perturbations were randomly distributed in each session. Participants wore a tethered safety harness to prevent injury during perturbation response (Fig. \ref{fig:fig3}A). A trip perturbation was induced by an abrupt deceleration \cite{sessoms2014method}, during which the right treadmill belt decelerated at a rate of -3.0 {m/s}$^2$ for 0.5 seconds. Conversely, a slip perturbation was induced by an abrupt acceleration in the opposite direction at 3.0 m/s$^2$ for 0.5 seconds, such that the right belt reached -1.5 m/s (Fig. \ref{fig:fig3}B). Perturbations occurred only on the right treadmill belt because gait phase was calculated based on the right heel strike. Following each perturbation, the right treadmill belt gradually (0.5 m/s$^2$) returned to the original treadmill speed (1.25 m/s). Onset timing and perturbation type were controlled and randomized using a program that remotely operated the treadmill and motion capture system (Lock Lab, Vicon, UK). This approach minimized the influence of anticipation on subjects' responses to perturbations and ensured even distribution of perturbation across the gait cycle (Fig. \ref{fig:fig3}C). Each perturbation was induced no earlier than 15 seconds into each trial, ensuring that the initial mean and standard deviation of each state were calculated from at least 10 steady-state cycles.

Participants wore a set of 8 reflective markers located on left and right side at the anterior and posterior superior iliac spine, heel, and 3rd metatarsal head. Motion capture marker data and ground reaction forces were processed (Nexus 2.10.1, Vicon, UK) and analyzed using a custom Python script (Python 12.3). Marker trajectories were filtered with the Plug-In-Gait Standard Woltring filter provided in the Nexus software. Trajectories with marker gaps of up to 25 frames were interpolated using a spline fill function. Using this processed data, \(\varphi\) was calculated using Eqs. 1 and 2. Heel strikes for each trial were identified using ground reaction forces from the right force plate \cite{hansen2002simple}. To simulate pseudo real-time performance, a moving average was calculated for each of the 16 kinematic states over the first 10 detected gait cycles of each trial. The mean and standard deviation were continuously updated using the preceding 10 gait cycles to account for natural and gradual shifts in gait patterns. A perturbation was identified when \(\varphi\) exceeded the optimized threshold. The percentage of the gait cycle elapsed between the known perturbation onset and the detection time was then calculated. Data from four trials were excluded from the analysis due to data corruption.

\section{Results}
\subsection{Optimal Perturbation Detection Threshold}
\begin{figure}
    \centering
    \includegraphics[width=1\linewidth]{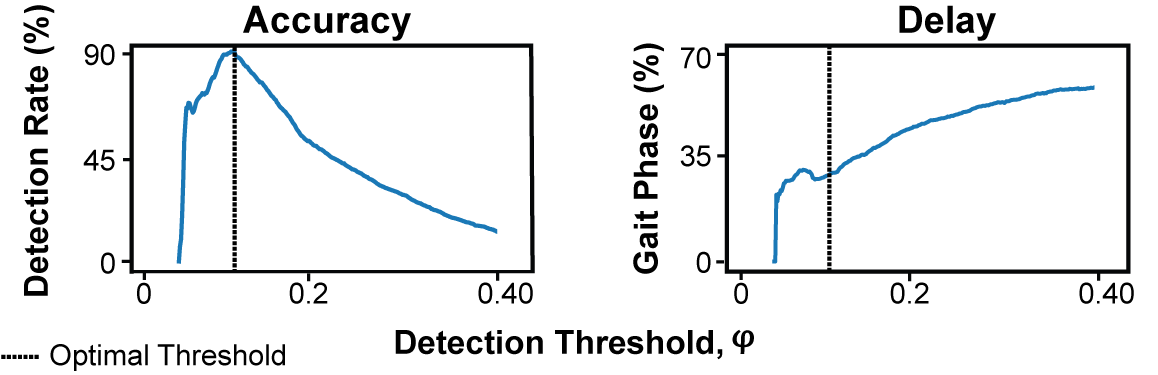}
    \caption{Parameter sweep result to determine an optimal perturbation detection threshold, \(\varphi\). A threshold value of 0.125 (black dashed lines) was found to be the optimal value.}
    \label{fig:fig4}
\end{figure}

A threshold for \(\varphi\) of 0.125 yielded the highest model accuracy of 87.65$\pm$0.61\% and the smallest detection delay of 28.12$\pm$0.51\% of the gait cycle (Fig. \ref{fig:fig4}). The upper limit of 87.65\% accuracy is likely due to instances where subjects exhibited a minimal physical response to the perturbation. Our post-hoc analysis found that the kinematic states often remained within their nominal trajectories in response to 5-cm magnitude perturbations that were induced in the direction of walking. 
\subsection{Perturbation Detection Performance on Pilot Data }
\begin{figure}[!t]
    \centering
    \includegraphics[width=1\linewidth]{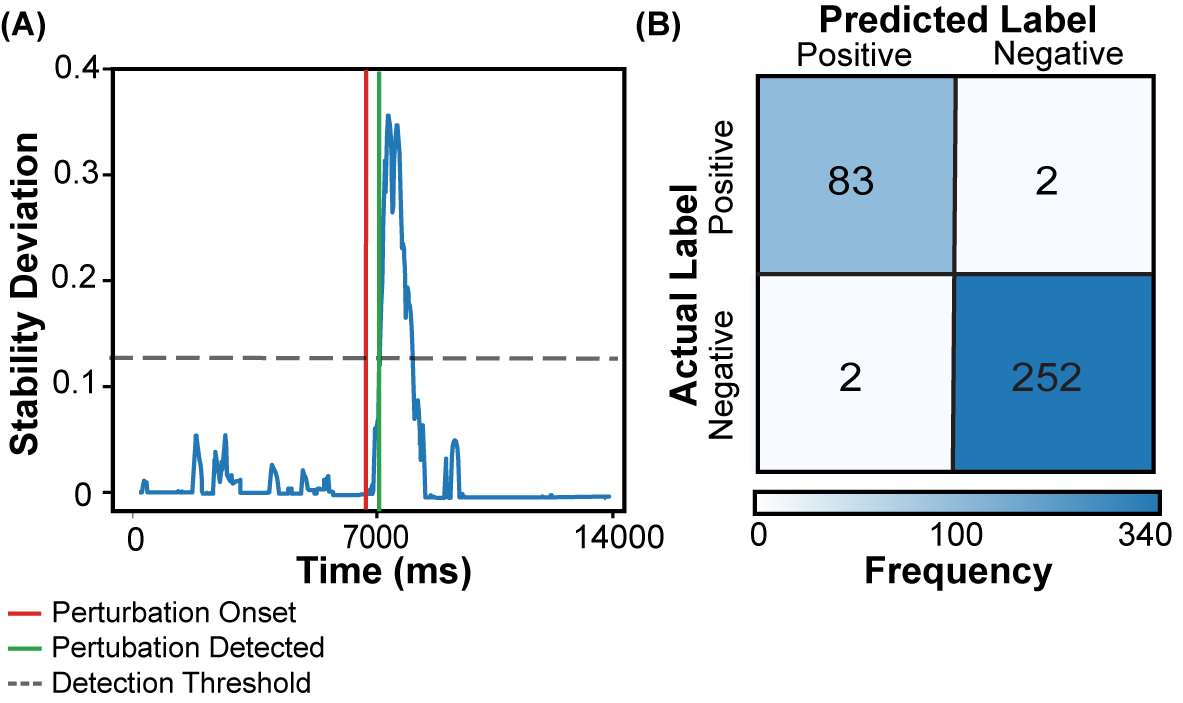}
    \caption{(A) A representative trial of our model performance in real-time. The red line represents the perturbation onset and the green line represents the time when our model detects the perturbation (detection threshold shown with the dashed grey line). (B) A confusion matrix summarizing our model performance. Positive and negative labels indicate perturbed and unperturbed gait cycles, respectively.}
    \label{fig:fig5}
\end{figure}

\begin{figure}[!b]
    \centering
    \includegraphics[width=1
    \linewidth]{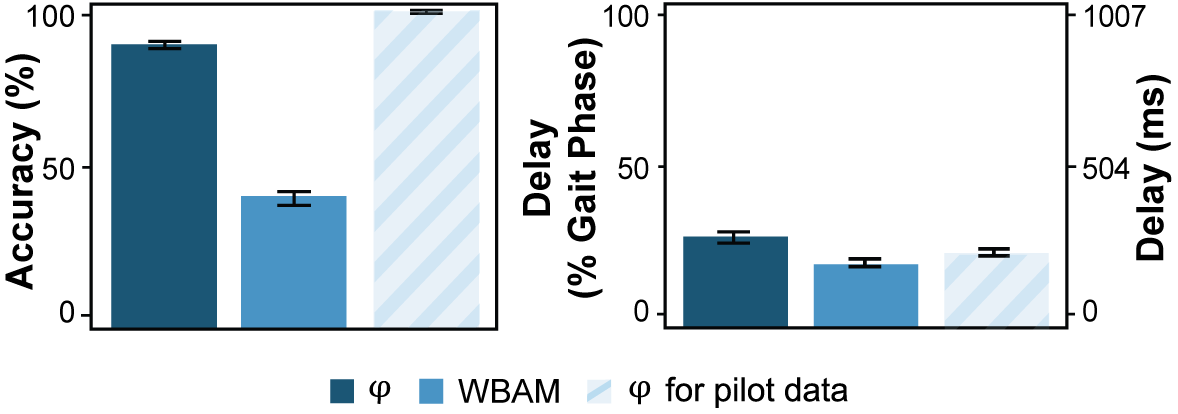}
    \caption{Perturbation detection performance result using our model (dark) and the WBAM-based benchmark (light) on accuracy (left) and delay (right). Solid bars indicate results on the open-source dataset and dashed bars represent results on our pilot experimental data. Error bars indicate $\pm$1 standard deviation.}
    \label{fig:fig6}
\end{figure}

The performance of our kinematic state-based detector was evaluated on the pilot experimental dataset using the same optimized detection threshold (Fig. \ref{fig:fig5}A). The model classified gait cycles in the pilot dataset as either perturbed or unperturbed with 98.8\% accuracy (Fig. \ref{fig:fig5}B). In addition, perturbations were detected 5.07\% of the gait cycle faster compared to the results in open-source dataset (Fig. \ref{fig:fig6}). 

\subsection{Baseline Model Comparison}
Implementation of the baseline WBAM-based model performed with 41.0\% less accuracy than our method for the open-source dataset. However, it resulted in a 11.11\% faster detection within the gait cycle. The WBAM-based results were obtained by taking the average across sagittal and frontal planes to provide a representative analysis (Fig. \ref{fig:fig6}) \cite{martelli2011detecting}. 

\section{Discussion}
In this paper, we proposed a novel perturbation detection model, which utilizes 16 lower-limb kinematic states and their deviations from steady-state gait dynamics. We established a local coordinate system to track relative positions to avoid misinterpreting natural drift in the state during walking. Using a data-driven approach, we further optimized a single threshold value to detect ground perturbations, \(\varphi\). We hypothesized that our kinematic state-based detection model would outperform the WBAM-based model under varying perturbation conditions across subjects. To test this hypothesis, we evaluated the performance of our model and the WBAM-based benchmark in detecting trip- and slip-type perturbations. Our study results corroborate our initial hypothesis that the ground perturbation detection model using kinematic states detected perturbations more accurately than the WBAM-based model. In the open-source dataset analysis, our model showed a 41\% improvement in detection accuracy compared to that of the WBAM-based model (Fig. \ref{fig:fig6}). However, contrary to our initial hypothesis, the WBAM-based model outperformed our model in the detection delay, demonstrating a faster response on average by 11\% earlier within the gait cycle. However, it is important to note that this delay was only calculated in instances of successful perturbation detections and may not be representative of the remaining 54\% of perturbations that the WBAM-based model failed to detect. Furthermore, since this was an offline analysis, it did not consider the time required to calculate WBAM. 

In our extended perturbation pilot experiment, our model demonstrated an improvement in detection accuracy compared to its performance on the open-source dataset, achieving 47.7\% higher accuracy than the WBAM-based model. Furthermore, our model detected 5\% faster within the gait cycle compared to its performance on the open-source dataset. The generalizability of our model across datasets showcased the potential benefit of integrating our approach into lower-limb exoskeleton control. The average stride duration across subjects in our pilot experiment was 1006.9$\pm$8.3 ms. Biomechanics literature states that older adults can require up to 469 ms to complete a stabilizing step in response to perturbations. Given that the detection timing of our model was 231 ms, our result indicated that the detection was early enough for an exoskeleton to respond (with assistance) within a single stride \cite{hardwick2022age, 7281223, JKPT2021}. Moreover, the generalizability of our model and its consistent performance across subjects showed that the system can potentially be deployed to exoskeletons with minimal user-specific tuning \cite{Kang2022}. This subject-independent characteristic of our model was highlighted with our consistent result across two distinct datasets.

While our model was capable of operating in real-time, gap fill and minimizing computation (though minimal compared to the WBAM approach) when using live-stream marker data needs further refinement for a real-time deployment. In future work, we aim to make our model fully autonomous by tracking kinematic states using on-board wearable sensors (e.g,. inertial measurement units) from an exoskeleton. Furthermore, we plan to reduce the amount of gait cycle data required to determine the initial nominal state trajectories (current system requiring 10 gait cycles), allowing for more dynamic locomotion \cite{Kim2024}. Current gait literature states that 2-3 strides are sufficient to capture the natural variance of healthy gait \cite{Mann1979}, however, further analysis is required to determine the minimum number of strides for our model application, specifically within older adult populations. 

Lastly, we plan to extend our model to determine both the direction and magnitude of each perturbation. The instantaneous sign and magnitude of each \(\alpha\) value can provide an estimate of the direction in which the individual’s feet and COM were perturbed as well as the relative magnitude of the perturbation (Fig. \ref{fig:fig3}D). We will evaluate the accuracy of these estimations and explore their potential to enhance the precision of perturbation responses in lower-limb exoskeletons. Moreover, we will evaluate our model under diverse ground perturbation conditions and subject demographics to improve the generalizability of our model. Although our model was trained on healthy young subjects, its performance at the same walking speed is likely comparable in healthy older adults. Given their longer reaction times, our method could be especially valuable for assessing reactive balance in older adults \cite{Coelho2023PosturalAdaptation}. However, additional data collection from older adult subjects is needed to evaluate the model's performance in this target population.

\section{Conclusion}
We developed a novel perturbation detection model by tracking lower-limb kinematic states and their deviations from steady-state gait dynamics during walking. Our model employed a weighted sum of state deviations and used a data-driven threshold to detect ground perturbations. In a pilot human subject experiment, our model achieved 98\% detection accuracy with a delay of 23\% within a gait cycle after a perturbation onset, outperforming the benchmark WBAM-based model. These results highlight the potential of our model to enable lower-limb exoskeletons to provide fast and precise balance support, particularly for older adult users.
\bibliographystyle{ieeetr}
\bibliography{Bibliography}
\end{document}